\documentclass[11pt,a4paper]{article}
\usepackage[hyperref]{emnlp2020}
\usepackage{times}
\usepackage{latexsym}

\usepackage{microtype}

\PassOptionsToPackage{dvipsnames}{xcolor}

\usepackage{acronym}
\usepackage{booktabs}
\usepackage[inline]{enumitem}
\usepackage{natbib}
\usepackage[dvipsnames]{xcolor}
\usepackage{times}
\usepackage{soul}
\usepackage{url}
\usepackage[utf8]{inputenc}
\usepackage[skip=0pt]{caption}
\usepackage{graphicx}
\usepackage{amsmath}
\usepackage{amsthm}
\usepackage{amssymb}
\usepackage{booktabs}
\usepackage{algorithm}
\usepackage{algorithmic}
\usepackage{changes}
\usepackage{tabularx}

\usepackage[subtle, wordspacing=normal, tracking=normal, bibnotes, charwidths=normal, leading, indent=normal, lists=normal, paragraphs=normal, mathspacing=normal, bibliography=normal]{savetrees}
\newcommand{\DIMEN}{\text{DIMEN}}
\newcommand{\distinct}{\text{distinct}}
\newcommand{\gram}{\text{-gram}}
\newcommand{\udimen}{u-DIMEN}
\newcommand{\ldimen}{l-DIMEN}
\newcommand{\hdimen}{hist^{\text{DI}}}
\newcommand{\wltwo}{\text{WL2}}
\newcommand{\ce}{\text{CE}}
\newcommand{\fl}{\text{FL}}
\newcommand{\tfl}{\text{TFL}}
\newcommand{\ldr}{\text{LDR}}
\newcommand{\tldr}{\text{TLDR}}
\newcommand{\down}{$\downarrow$}
\newcommand{\up}{$\uparrow$}
\newcommand{\cosw}{\text{cosw}}

\acrodef{Seq2Seq}{sequence-to-sequence}
\acrodef{MT}{machine translation}
\acrodef{NMT}{neural machine translation}
\acrodef{DIMEN}{DIversity Metric based on N-grams}
\acrodef{RNN}{recurrent neural network}
\acrodef{LM}{language modeling}
\acrodef{NLG}{Natural Language Generation}
\acrodef{NLP}{Natural Language Processing}
\acrodef{QA}{question answering}
\acrodef{TLDR}{token loss dynamic reweighting}
\acrodef{LDR}{loss dynamic reweighting}
\acrodef{CE}{cross-entropy}
\acrodef{FL}{Focal Loss}
\acrodef{TFL}{token-level Focal Loss}
\acrodef{IF}{Input-feeding}

\newcommand{\OurMethod}{\ac{TLDR}}

\AtBeginDocument{%
  \providecommand\BibTeX{{%
    \normalfont B\kern-0.5em{\scshape i\kern-0.25em b}\kern-0.8em\TeX}}}

\aclfinalcopy 

\newcommand\BibTeX{B\textsc{ib}\TeX}

\title{TLDR: Token Loss Dynamic Reweighting \\ for Reducing Repetitive Utterance Generation}

\author{
  Shaojie Jiang$^1$ \quad Thomas Wolf$^2$ \quad Christof Monz$^1$ \quad Maarten de Rijke$^1$  \\
  $^1$University of Amsterdam, Amsterdam, The Netherlands \\
  $^2$Hugging Face, Brooklyn, New York, United States \\
  \{s.jiang, c.monz, m.derijke\}@uva.nl, thomas@huggingface.co
}

\date{}

\begin{document}
\maketitle
\begin{abstract}
  \ac{NLG} models are prone to generating repetitive utterances.
  In this work, we study the repetition problem for encoder-decoder models, using both \ac{RNN} and transformer architectures.
  To this end, we consider the chit-chat task, where the problem is more prominent than in other tasks that need encoder-decoder architectures.
  We first study the influence of model architectures.
  By using pre-attention and highway connections for \acp{RNN}, we manage to achieve lower repetition rates.
  However, this method does not generalize to other models such as transformers.
  We hypothesize that the deeper reason is that in the training corpora, there are \emph{hard tokens} that are more difficult for a generative model to learn than others and, once learning has finished, hard tokens are still under-learned, so that repetitive generations are more likely to happen.
  Based on this hypothesis, we propose \acfi{TLDR}\acused{TLDR} that applies differentiable weights to individual token losses.
  By using higher weights for hard tokens and lower weights for easy tokens, \ac{NLG} models are able to learn individual tokens at different paces.
  Experiments on chit-chat benchmark datasets show that \OurMethod{} is more effective in repetition reduction for both \ac{RNN} and transformer architectures than baselines using different weighting functions.
\end{abstract}


\section{Introduction}

It has been widely reported that \acf{NLG} models are prone to generating repetitive utterances, which is a cross-task and cross-architecture problem.
E.g., the repetition problem has been observed in \ac{QA}~\citep{fan2019eli5}, \ac{LM}~\citep{holtzman2019curious, keskar2019ctrl, welleck2019neural}, abstractive summarization~\citep{suzuki2017cutting,nallapati2016abstractive}, \ac{MT}~\citep{tu2016modeling, mi2016coverage, tu2017context}, and image captioning~\citep{cornia2018paying}, etc.
The problem arises with most popular neural architectures, including \ac{Seq2Seq} modeling using \acfp{RNN} \citep{sutskever2014sequence} and transformers \citep{vaswani2017attention}.
Previous approaches to the repetition problem have mostly been developed for decoder-only models \citep{holtzman2019curious, keskar2019ctrl, welleck2019neural}.
Whenever it is possible to use them, encoder-decoder models usually outperform decoder-only models \citep{raffel2019exploring}; because of this, we study the repetition problem for encoder-decoder models in this work.
We base our experiments on the chit-chat task,\footnote{Throughout this paper, we use `chit-chat,' `open-domain dialogue,' and `response generation' to refer to the same task interchangeably, and use `chatbot' to refer to the model trained for this task.} where the repetition problem is more prominent than other tasks that need encoder-decoder architectures.
Figure~\ref{fig:rep_ex} is an illustration of a chit-chat scenario.
%
\begin{figure}[h]
  \centering
  \fontfamily{cr}\selectfont
  \footnotesize
  \noindent%
  \begin{tabularx}{\textwidth}{@{}p{.3\linewidth}p{.65\linewidth}}
    User Message: &Hi, Jim. How are you? I haven't seen you for a while.\\
    \raggedright System Response: &i've been out of town. i've been out of town. i've been out of town and got some old ideas on i've been out of town.
  \end{tabularx}
  \caption{An illustration of the repetitive response problem.}
  \label{fig:rep_ex}
\end{figure}
The repetition problem on which we focus in this paper is \emph{not} to be confused with the problem where responses of different turns share a great deal of similarity, which is sometimes also referred to as repetitive responses~\citep{li2016deep}.

We first surmise that the repetition problem is caused by model architectures.
By studying existing approaches to repetition reduction involving changing model architectures, we identify that a certain model design for \acp{RNN} can help reduce repetition.
More specifically, by using pre-attention (\S \ref{section:hw-s2s}) together with \acp{RNN} instead of the conventional usage of post-attention, the model generates less repetitive utterances.
The usage of highway connection \citep{srivastava2015training} after pre-attention can further reduce repetition.
Though it is effective for \acp{RNN}, this architectural design unfortunately does not generalize to transformers.
Therefore, we believe that there are deeper reasons to the repetition problem.

Inspired by the well-studied class imbalance problem \citep{lin2017focal}, we hypothesize that in training corpora there are \emph{hard} tokens and \emph{easy} tokens; if the training stops when hard tokens are still under-learned, repetitive generations are more likely to happen.
We empirically validate this hypothesis by using the \ac{FL} method proposed by \citet{lin2017focal}.
Based on this finding, we propose \acfi{TLDR}, which is inspired by \ac{FL} (\S\ref{section:tldr}) and which assigns differentiable weights to training losses according to the training difficulty of tokens.
By dynamically using higher weights for hard tokens and lower weights for easy tokens, the model is able to learn tokens at different paces.
Experiments on benchmark chit-chat datasets show that \OurMethod{} is more effective in reducing repetitions for \ac{RNN} and transformer architectures than the baselines.

The main contributions of this work include:
\begin{itemize}[leftmargin=*,nosep]
\item We discover that \acp{RNN} with pre-attention and highway connections can help reduce repetition.
\item We hypothesize that the deeper reason for repetitions is due to hard tokens, and empirically validate this by using FL.
\item We propose a more effective \acf{TLDR} method for reducing repetitive generations that is architecture insensitive.
\item We share our source code at \href{https://github.com/ShaojieJiang/tldr}{https://github.com/ ShaojieJiang/tldr}.
\end{itemize}



\section{Related Work}
\label{section:related work}

\subsection{Repetition reduction}

There are generally three types of approach to the repetition problem, namely architecture-based methods, auxiliary loss, and sampling-based methods.

\textbf{Architecture-based methods} try to address the repetition problem by exerting inductive biases on model architectures, such as introducing coverage vectors to attention mechanisms \citep{tu2016modeling, mi2016coverage}.
However, these solutions are based on the desirable property of source-target correspondence of \ac{NMT} tasks, which rarely holds for other tasks like open-domain dialogue generation and abstractive summarization.
Instead, we observe that by using pre-attention (\S\ref{section:hw-s2s}), \ac{RNN} models can generate less repetitive responses.
Adding a highway connection to attention layers can further reduce repetition.
However, it is not yet clear which modifications to transformers can help reduce repetitive generations, which we surmise to be mainly due to their lack of temporal inner states.

\textbf{Auxiliary loss} adds an unlikelihood item of repeated tokens to the conventional \ac{CE} loss \citep{welleck2019neural, li2019don} to discourage repetitive generations.
However, besides the need for extra data types \citep{li2019don} or extra processing on model generations \citep{welleck2019neural}, it is also worth noting that the repetition problem seldom exists in the training data (\S\ref{section:results}), and therefore, the models are \emph{not} trained to be repetitive in the first place.
Our experiments show that the repetition problem is mainly due to the fact that trained models have different levels of proficiency toward easy and hard tokens.
Based on this, we propose to weight the token losses w.r.t.\ their training difficulties to address the repetition problem.
Though, normally, hard tokens already result in larger gradients, apparently this is not enough to learn them well as their effects may be dissolved by large amount of easy tokens.

\textbf{Sampling based approaches} discourage repetition at the sampling stage, e.g., by creating an n-gram blacklist \citep{paulus2017deep}.
Some more sophisticated methods include discounting the scores of previously generated tokens \citep{keskar2019ctrl}, top-$k$ random sampling \citep{fan2018hierarchical}, and top-$p$ sampling with a dynamic nucleus \citep{holtzman2019curious}.
These methods can only be used at inference time to remedy the repetition problem.
In this work, we tackle the problem at a deeper level and show that by dynamically reweighting token losses at training time, the problem can be largely alleviated.

\subsection{Class imbalance}

Class imbalance is a common cause of poor performance of trained models.
To mitigate the low efficiency of heuristic sampling approaches to balancing classes, \citet{lin2017focal} try to balance the \ac{CE} loss by weighting easy and hard examples differently for visual object detection.
Similarly, FACE \citep{jiang2019improving} tries to balance the \ac{CE} loss for frequent and rare tokens through heuristics such as token frequency for improving generation diversity.
In our case, however, the frequency of a token does not necessarily reflect its training difficulty, as the difficulty is also influenced by the context of this token.

We hypothesize that the repetition problem is also caused by different training difficulties of tokens.
This hypothesis is in line with \citep{fadaee2018back}, where the authors improve the learning of hard tokens by using back-translation and by sampling examples with difficult words in similar contexts.
Considering that at the token level in the chit-chat task, balancing easy and hard tokens through sampling is infeasible, loss reweighting such as Focal Loss \citep{lin2017focal} is a preferable solution.
Focal Loss down-weights\footnote{Models trained without weighting can be thought of as using a uniform weight of 1.} both easy and hard examples, albeit at different ratios, which is likely to slow down training.
Instead, we propose a different weighting function that simultaneously up-weights hard tokens and down-weights easy tokens.



\section{RNN with Pre-Attention and Highway Connection} 
\label{section:hw-s2s}

In this section, we introduce some architectural modifications that alleviate the repetition problem.

\subsection{Notation and response generation}
\label{section:notations}

We write $x = (x_1, x_2, \dots, x_{|x|})$, $y = (y_1$, $y_2$, \dots, $y_{|y|})$ for message and response utterances, respectively.
Given $x$, the objective of a chatbot model parameterized by $\theta$ is to assign a higher conditional probability $p(y|x; \theta)$ to the ground truth response $y$ than to other responses, which usually follows a sequential decomposition:
\begin{equation}
  \label{eq:objective}
  p(y|x; \theta) = \prod_{t=1}^{|y|} p(y_t|y_{<t}, x; \theta).
\end{equation}
Here, $y_{<t} = (y_1, y_2, \dots, y_{t-1})$, and $p(y_t|y_{<t}, x; \theta)$ represents the model-predicted probability for token $y_t$.

For the sake of clarity, we further denote
$p_t = p(y_t|y_{<t}, x; \theta)$.
Training the above chatbot model is usually achieved by minimizing the \ac{CE} loss:
\begin{equation}
  \label{eq:ce}
  \ce(p_t) = -\log(p_t).
\end{equation}
For estimating $p_t$ using an \ac{RNN} model, we largely follow the notation in \citep{bahdanau2015neural} by denoting the decoder output and attention output at decoding step $t$ as $s_t$ and $c_t$, respectively.
Readers are referred to \citep{bahdanau2015neural} for more details.

\subsection{Pre-attention}

\begin{figure*}
  \centering
  \includegraphics[width=.75\textwidth]{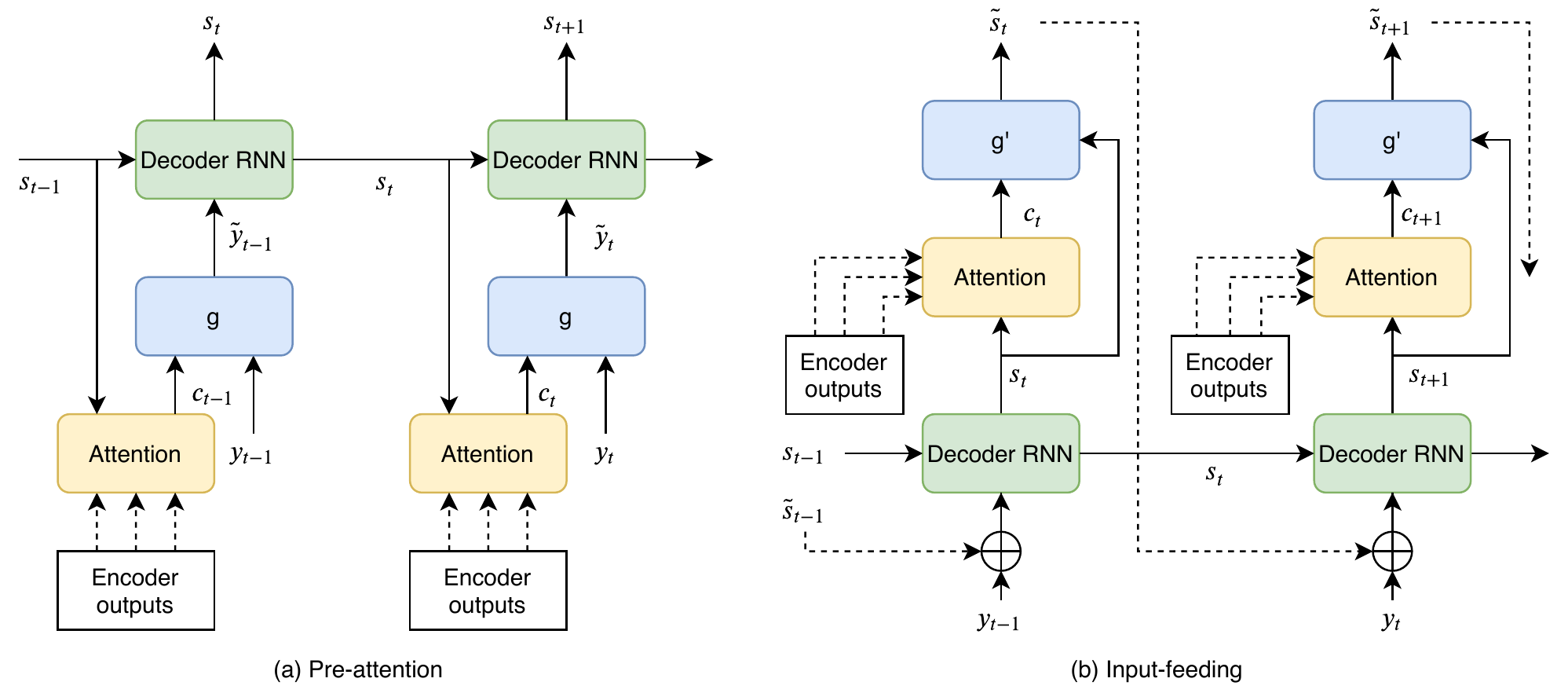}
  \caption{Pre-attention vs \ac{IF} attention. $\oplus$ denotes vector concatenation.}
  \label{fig:if}
\end{figure*}

Inspired by \ac{IF} attention \citep{luong2015effective}, we find that pre-attention performs better w.r.t.\ repetition rates.
Figure~\ref{fig:if} is an illustration of the difference between pre-attention and \ac{IF} attention.

Formally, the input of \ac{RNN} using \ac{IF} attention at each decoding step $t$, i.e., $y_{t-1}$, is modified:
\begin{equation}
  \tilde{y}'_{t-1} = [y_{t-1}; \tilde{s}_{t-1}],
\end{equation}
with $[\cdot~; \cdot]$ denoting vector concatenation.
Besides, at step $t$ the decoder output $s_t$ is also transformed with attention output $c_t$:
\begin{equation}
  \tilde{s}_t = g'(c_t, s_t).
\end{equation}
The $g'(\cdot)$ function used in \ac{IF} attention is a vector concatenation followed by a linear projection, to resize $\tilde{s}_t$ to the dimension of $s_t$.

We find that only transforming the input $y_{t-1}$ is more helpful for reducing repetition, which leads us to pre-attention:
\begin{equation}
  \tilde{y}_{t-1} = g(c_{t-1}, y_{t-1}),
\end{equation}
with $g(\cdot)$ being a transformation function, and $\tilde{y}_{t-1}$ has the same dimensionality as $y_{t-1}$.
The $g(\cdot)$ function can take the form of $g'(\cdot)$, but in this paper, we observe that by using a highway connection for $g(\cdot)$, the repetition rate can be further reduced.
The highway connection is formulated as follows:
\begin{align}
  \label{eq:highway}
  \tilde{y}_{t-1} &= \textit{Highway}(c_{t-1}, y_{t-1}) \nonumber\\
  &= z \circ c_{t-1} + (1 - z) \circ y_{t-1},
\end{align}
where $\circ$ represents element-wise product; $z$ is a learnable gating vector, through which the model learns how to effectively combine attentional information $c_{t-1}$ and input information $y_{t-1}$.
Readers are referred to \citep{srivastava2015training} for details of the highway connection.

\subsection{Discussion}

The attention mechanism does not have internal states to keep track of attentions previously paid, therefore when using post-attention, similar attentions may be paid repeatedly, increasing the chance of repetition.
On the other hand, when using pre-attention as part of the input to \acp{RNN}, it enables \ac{RNN} states to keep track of previous attention status.
Furthermore, by using highway connections, the model learns to more effectively combine context and step-wise input than simple concatenation and linear projection.

However, noticing that the transformer model \citep{vaswani2017attention} already has an analogous architecture as our pre-attention, we tried to apply highway connections to the transformer attention outputs, but have not yet been successful with reducing repetitive generations in this manner.
We suspect that this is due to the lack of temporal inner states in transformers, and more work needs to be done to mitigate this drawback in order to reduce repetition.
Meanwhile, we believe there are more fundamental reasons for the repetition problem.



\section{Token Loss Dynamic Reweighting}
\label{section:tldr}

Inspired by \citet{lin2017focal}, we hypothesize that the difficulty of some training examples is the main reason for repetition.
Based on this hypothesis, example reweighting methods such as \ac{FL} \citep{lin2017focal} can be used to alleviate the repetition problem.
In this section, we first adapt FL to our chit-chat task, and later we present a more effective weighting method than FL.

\subsection{Example-level loss reweighting}
Using the notation introduced in \S \ref{section:notations} and writing $p^y$ for $p(y|x; \theta)$, a direct application of \ac{FL} to the response generation task can be formulated as follows:
\begin{equation}
  \label{eq:fl}
  \fl(p^y) = w \cdot L(y).
\end{equation}
Here, $w = (1 - p^y)^\gamma$ is the weighting coefficient, which is scaled by an exponential focusing parameter $\gamma \geq 0$, with $(1 - p^y)$ reflecting the training difficulty of an example, that is, a pair $(x, y)$.
$L(y)$ denotes the loss of predicting $y$ that is usually calculated as an average over the \ac{CE} losses of all tokens that make up $y$:
\begin{equation}
  L(y) = \frac{1}{|y|} \sum_{t=1}^{|y|} \ce(p_t).
\end{equation}
Since $p_t \in [0, 1]$, in practice $p^y$ is usually extremely small according to the sequential decomposition Eq.~\eqref{eq:objective}, turning almost all training examples into hard examples.
Therefore, we propose to approximate the training difficulty of $(x, y)$ w.r.t. the average of $p_t$ as follows:
\begin{equation}
  \tilde{p}^y = 1 - \frac{1}{|y|} \sum_{t=1}^{|y|} p_t,
\end{equation}
which is more suitable for our task.

However, there is another unwanted feature of FL.
Although $w$ in Eq. \eqref{eq:fl} is higher for hard examples than for easy ones, to make the model \emph{focus} on examples that are hard to train, which is essentially what the method is named after, this method is likely to slow down the training as $w$ is always smaller than 1.
To address this, we propose to use a different weighting function:
\begin{equation}
  \label{eq:cosw}
  \cosw(\tilde{p}^y) = \cos(\tilde{p}^y * \pi) + 1,
\end{equation}
which projects all probabilities to the weight domain of $[0, 2]$.
With $\tilde{p}^y \leq 0.5$ indicating a hard example, which results in $\cosw(\tilde{p}^y) > 1$, we up-weight the loss of hard examples, and otherwise we down-weight easy examples with $\tilde{p}^y > 0.5$.
To distinguish our method (using \eqref{eq:cosw}) from the adapted \ac{FL}, we refer to it method as \acfi{LDR}\acused{LDR}:
\begin{equation}
  \label{eq:ldr}
  \ldr(p^y) = \cosw(\tilde{p}^y) \cdot L(y).
\end{equation}

\subsection{Token-level loss reweighting}

Noticing from Eq. \eqref{eq:ce} that in our task, \ac{CE} losses are calculated for each token, we can alternatively infer the difficulty at the token-level and apply FL or \ac{LDR} methods thereafter.
To achieve this, we use $p_t$ instead of $p^y$ for Eq. \eqref{eq:fl}, resulting in \acfi{TFL}:
\begin{equation}
  \label{eq:tfl}
  \tfl(p_t) = w_t \cdot \ce(p_t),
\end{equation}
where $w_t = (1 - p_t)^\gamma$.
Using $p_t$ instead of $\tilde{p}^y$ and $p^y$ for Eq.~\eqref{eq:ldr} results in \acfi{TLDR}:
\begin{equation}
  \label{eq:tldr}
  \tldr(p_t) = \cosw(p_t) \cdot \ce(p_t).
\end{equation}

Although treating each token as an example is somewhat counterintuitive, it has been shown in \citep{jiang2019improving} that such a token-level formulation can have a more direct impact on the generation of each token.
We confirm in \S \ref{section:results} that \ac{TLDR} results in less repetition than \ac{LDR}.

\subsection{Gradient analysis}

We now compare the gradients of our loss function TLDR to those of CE and TFL w.r.t. $p_t$.
The gradient function of CE is:
\begin{equation}
  \frac{\partial \ce(p_t)}{\partial p_t} = -\frac{1}{p_t}.
\end{equation}
The gradient function of TFL is:
\begin{equation}
\begin{split}
  &\frac{\partial \tfl(p_t)}{\partial p_t} = \\
   & \gamma (1 - p_t)^{\gamma - 1} \cdot \log(p_t) - \frac{(1 - p_t)^\gamma}{p_t}.
\end{split}  
\end{equation}
Finally, the gradient function of TLDR is:
\begin{equation}
  \begin{aligned}
    &\frac{\partial \tldr(p_t)}{\partial p_t} = \\
    &-\pi \sin(p_t \cdot \pi) \log(p_t) - \frac{\cos(p_t \cdot \pi) + 1}{p_t}.
  \end{aligned}
\end{equation}

\begin{figure}
  \centering
  \includegraphics[width=.9\linewidth]{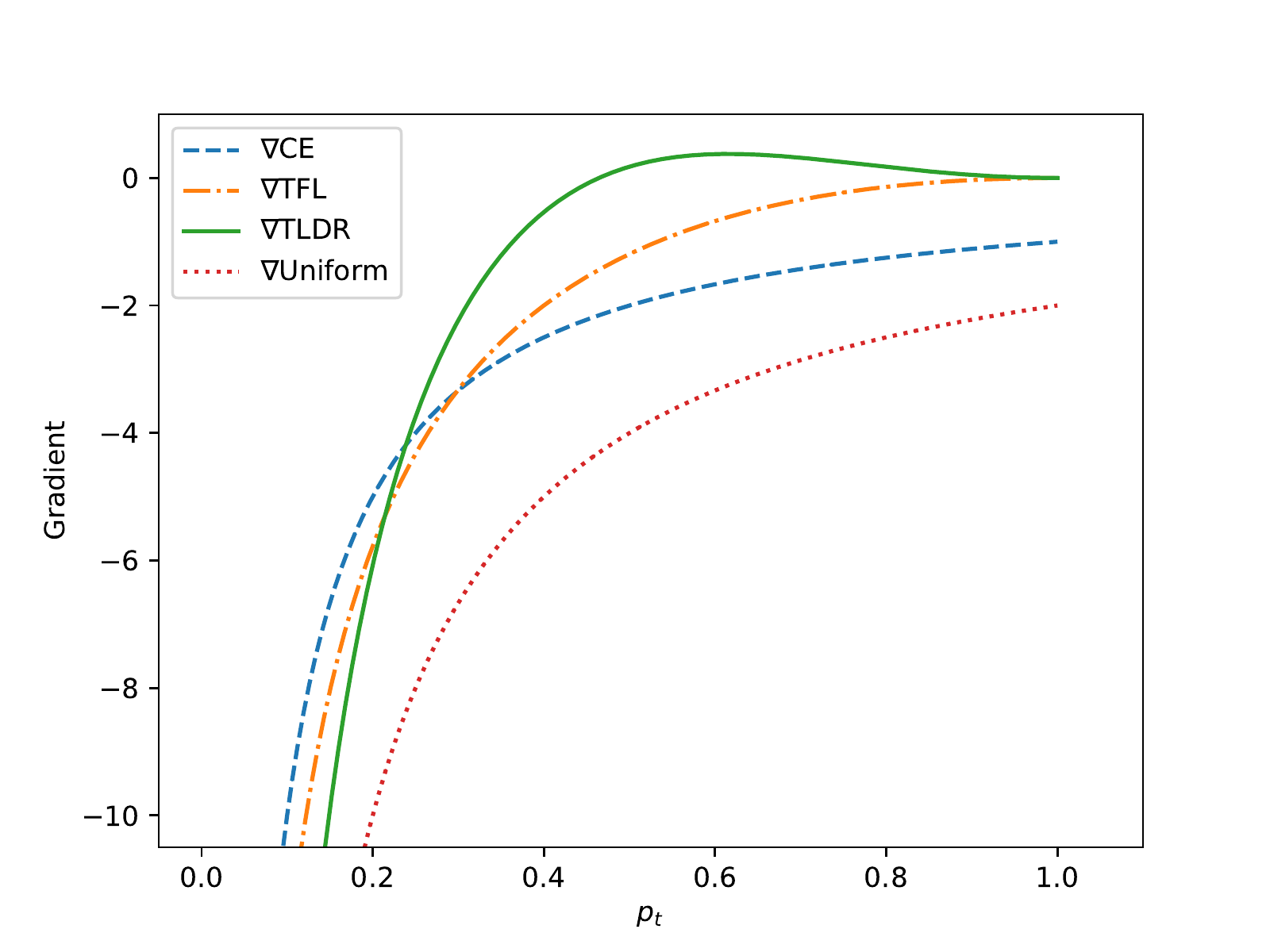}
  \caption{Gradient curves of each loss function.
  For TFL, we use $\gamma = 2$ as recommended by \citep{lin2017focal}.}
  \label{fig:gradients}
\end{figure}

\noindent%
To see the differences between the three loss functions more clearly, we visualize their curves in Figure~\ref{fig:gradients}.
Compared to CE and TFL, the gradients for easy tokens derived by TLDR are suppressed and gradients for hard tokens are amplified.



\section{Experimental Setup}
\label{section:experimental setup}

For our experiments, we use 3 multi-turn chit-chat datasets: Dailydialog \citep{li2017dailydialog}, Cornell Movie Dialogs \citep{danescu2011chameleons} and Ubuntu Dialogue \citep{lowe2015ubuntu}, in increasing order of sizes.
As a quantitative measure of system generated responses, we use BLEU-4 scores \citep{papineni2002bleu}.
For repetition measurement, unlike existing work utilizing repeated n-gram counts \citep{mi2016coverage} or n-gram diversity \citep{li2016diversity, welleck2019neural} that reflect only one order of n-gram repetition (usually 4-gram), we propose to use a \acf{DIMEN}, which is inspired by BLEU by weighted-averaging the n-gram diversities at multiple granularities, thus being more comprehensive.
To introduce \ac{DIMEN}, we first introduce the definition of n-gram diversity proposed by \citet{li2016diversity}:
\begin{equation}
\begin{split}
  \label{eq:distinct}
  &\distinct(text, k) = \\
   &\mbox{}\quad\frac{|\{k\gram(text) \}|}{\max(|1\gram(text)| - k, 1)}.
\end{split}   
\end{equation}
Here, $k\gram(text)$ means getting the n-gram list of $text$ for $n=k$;
$\{\cdot\}$ maps a list to a set, and $|\cdot|$ denotes the cardinality of a set/list, respectively.
For $text$ with less than $k$ unigrams, we clip the denominator to be 1 such that $\distinct(text, k) = 1$.
With n-gram diversity defined, the DIMEN score of a given $text$ is calculated as
\begin{equation}
  \label{eq:dimen}
  \mbox{}\hspace*{-2mm}
  \DIMEN(text, n) {=} \sum_{k=1}^n \alpha_k \cdot \distinct(text, k),
  \hspace*{-2mm}\mbox{}
\end{equation}
where $n$ is the highest order of n-grams and $\alpha_k$ is the weight coefficient corresponding to $k\gram$ with $\sum_1^n \alpha_k = 1$.

The $text$ argument can be either a response utterance, in which case we refer to the score as \udimen{}, or a list of response utterances for a whole validation set, where we refer to as \ldimen{}.
\udimen{} and \ldimen{} scores are compensating.
With $\DIMEN(\cdot) \in [0, 1]$, a high \udimen{} score represents a non-repetitive utterance, and a high \ldimen{} score represents a diverse list of utterances.

To further measure the repetition performance on a validation set, we can consider averaging the \udimen~scores of each utterance.
However, since the majority of system generated utterances are non-repetitive, the averaged \udimen~score can be very close to 1 and thus indistinguishable from model to model.
Instead, we first bin the \udimen~scores into $m$ buckets to create a histogram vector $\hdimen \in \mathbb{Z}^m$, and then calculate the weighted L2-norm 
of the count of each bucket to emphasize the overall repetition performance of a model.
Formally,
\begin{equation}
  \label{eq:wl2}
  \wltwo(\hdimen) = \sqrt{\sum_{i = 1}^{m} \beta_i \cdot (\hdimen_i)^2},
\end{equation}
where $\beta \in \mathbb{R}^m$ is a weighting vector.
When we assign higher weights to bins with lower \udimen{} scores, the \wltwo{} score emphasizes highly repetitive utterances.

With all the datasets and metrics introduced, we seek to answer the following research questions through our experiments:
\begin{enumerate}[label=(RQ\arabic*),leftmargin=*,nosep]
\item Are pre-attention and highway connection helpful for reducing repetition of RNN Seq2Seq?
\item Is our hypothesis on hard tokens correct?
\item How effectively can LDR and TLDR reduce repetition for both RNN and transformer models?
\item Is the improvement of LDR and TLDR merely due to up-weighting?
\end{enumerate}
To answer RQ1, we use an \ac{RNN} Seq2Seq with post-attention as the baseline.
To answer RQ2, we obtain results of using TFL on both RNN and transformer models.
For answering RQ3, we compare the results of LDR and TLDR to those of baseline models trained without weighting and with TFL weighting.
For RQ4, we also add a baseline using uniform weights, with $w = 2$ for all training examples.

We conduct our experiments on the ParlAI framework \cite{miller2017parlai}.
Below we introduce the model parameters.
Unless stated otherwise, we use the same settings for different models on each dataset whenever possible.
For \ac{RNN} \ac{Seq2Seq}, we use 2-layer LSTMs \citep{hochreiter1997long} with a hidden size of 512 and separate embedding matrices with an embedding size of 200 for both the encoder and the decoder.
We use the `general' attention variant from \citep{luong2015effective}.
For transformer \ac{Seq2Seq}, we use 6-layer transformer blocks with 8 attention heads \citep{vaswani2017attention} and an embedding size of 256 with separate matrices for both the encoder and the decoder.
We use 800 as the feed-forward layer size with `relu' activation.
We train both RNN and transformer Seq2Seq models using Adam optimizer \citep{kingma2014adam} with fixed learning rate of 0.001 and $\beta_1 = 0.9, \beta_2 = 0.999$.
The gradients of \acp{RNN} are clipped with L2-norm \citep{pascanu2013difficulty}, no larger than 5 during training, while for transformers we clip to 1.
For all the intermediate outputs of both models, we use dropout rate of 0.1 \citep{srivastava2014dropout}.

We train the models on the Dailydialog and Movie Dialogs datasets for 100 epochs and check the performances on the validation sets every 0.5 epochs.
On the Ubuntu Dialogue dataset, we train the models for 30 epochs with a validation interval of 0.1 epochs.
We save checkpoints after every validation and select the best checkpoint according to the lowest repetition rate.
Since the repetition rate can sometimes be very low, this checkpoint-saving strategy may fail occasionally, in which case we use the last checkpoint during training.
For all three datasets, we tokenize the utterances using the NLTK \citep{loper2002nltk} tokenizer and keep the most frequent 30,000 tokens according to their training sets, respectively.
We use up to 3 turns of history as the input message, and truncate the message to 128 tokens and response to 32 tokens.
A batch size of 256 is used.

For \ac{FL}, we use a focusing factor of $\gamma = 2$ as recommended in \citep{lin2017focal}.
For the hyper-parameters introduced in this work, we use $n = 4$ for calculating the DIMEN scores, with $\alpha = [0.25, 0.25, 0.25, 0.25]$.
We group the \udimen~scores into $m = 10$ bins, with the weighting vector $\beta = [0.9, 0.8, \dots, 0.0]$.



\section{Results}
\label{section:results}

\begin{table*}
  \centering
  \footnotesize
  \setlength{\tabcolsep}{4.3pt}
  \begin{tabular}{ll ccc l ccc l ccc}
    \toprule
    &&\multicolumn{3}{c}{Dailydialog} &&\multicolumn{3}{c}{Movie Dialogs} &&\multicolumn{3}{c}{Ubuntu Dialogue} \\
    \cmidrule{3-5} \cmidrule{7-9} \cmidrule{11-13} 
    &&WL2\down &BLEU\up &\ldimen\up &&WL2\down &BLEU\up &\ldimen\up &&WL2\down &BLEU\up & \ldimen\up \\
    \midrule
    (a) &RNN &8.04 &\textbf{19.63} &0.35 &&36.71 &0.35 &0.25 &&208.7 &0.08 &\textbf{0.10} \\
    (b) &RNN w/ pre-attn &5.70 &18.17 &\textbf{0.36} &&\textbf{28.15} &\textbf{0.41} &\textbf{0.27} &&173.0 &\textbf{0.09} &0.08 \\
    (c) &(b) w/ highway &\textbf{5.50} &19.60 &\textbf{0.36} &&29.00 &0.38 &0.25 &&\textbf{157.1} &0.07 &0.08 \\
    \midrule
    (d) &(b) w/ \ldr &5.20 &11.73 &0.36 &&26.90 &0.10 &0.22 &&230.4 &0.06 &0.06 \\
    (e) &(b) w/ \tfl &4.67 &11.48 &0.33 &&\textbf{19.11} &0.38 &0.26 &&153.3 &0.06 &\textbf{0.08} \\
    (f) &(b) w/ uniform  &5.32 &\textbf{18.47} &\textbf{0.37} &&31.96 &\textbf{0.43} &\textbf{0.27} &&132.0 &\textbf{0.08} &\textbf{0.08} \\
    (g) &(b) w/ \tldr &\textbf{2.71} &17.61 &0.36 &&22.45 &0.39 &0.26 &&\textbf{101.4} &0.05 &\textbf{0.08} \\
    \midrule
    (h) &transformer &7.72 &20.38 &0.35 &&42.50 &\textbf{0.57} &\textbf{0.30} &&79.84 &0.05 &0.11 \\
    (i) &(h) w/ \ldr &7.92 &19.19 &\textbf{0.36} &&42.40 &0.28 &0.29 &&255.0 &0.05 &0.12 \\
    (j) &(h) w/ \tfl &7.33 &20.69 &0.35 &&38.52 &0.51 &0.29 &&103.2 &0.05 &\textbf{0.15} \\
    (k) &(h) w/ uniform &\textbf{6.42} &\textbf{20.74} &\textbf{0.36} &&39.13 &0.56 &0.29 &&150.6 &\textbf{0.07} &\textbf{0.15} \\
    (l) &(h) w/ \tldr &6.67 &20.13 &0.35 &&\textbf{28.61} &0.53 &0.29 &&\textbf{75.11} &0.05 &0.13 \\
    \midrule
    &Human &4.78 &-- &0.43 &&40.24 &-- &0.42 &&32.01 &-- &0.42 \\
    \bottomrule
  \end{tabular}
  \caption{Results for all our methods and the baselines.
    `w/ pre-attn' and `w/ uniform' are short for `with pre-attention' and `with weight of 2 uniformly'.
    Best scores in each column within each row-wise block are highlighted in \textbf{bold face}.
    \up{}/\down{} indicate higher/lower is better.
    All results are on the test set.}
  \label{tab:results}
\end{table*}

The results of all models on the three datasets are shown in Table~\ref{tab:results}.
Since the validation repetition (\wltwo{} scores) of models can be very low in the early stage of training when the \ldimen{} score is still very low, we mainly use \ldimen{} to make sure that the results are shown for checkpoints at similar stages.

Table~\ref{tab:results} is divided into three main blocks row-wise.
The first block ranging from rows (a)--(c) is for answering RQ1.
The second (rows (d)--(g)) and third (rows (h)--(l)) blocks are for answering RQ2 and RQ3, with \acp{RNN} and transformer models, respectively.
For reference, we also include the \emph{Human} performance calculated using the ground truth of each dataset in the bottom row of Table~\ref{tab:results}.

\textbf{Answer to RQ1:}
Row (a) of Table \ref{tab:results} are the results for RNN models with post-attention.
From row (b), we can see that pre-attention is effective in reducing repetition on all three datasets.
On top of pre-attention, row (c) shows that highway connections are helpful with repetition reduction on two datasets, namely Dailydialog and Movie Dialogs.
BLEU scores show that the quality of the utterances generated by our models (b) and (c) are on a par with those generated by the baseline (a).
These results indicate that proper architectural modifications can indeed reduce repetition, and more work needs to be done to find such modifications for transformers.

\textbf{Answer to RQ2:}
From rows (e) and (j), we can see that in most cases \tfl{} helps to reduce repetition, compared to rows (b) and (h), respectively.
This empirically shows that our hypothesis on hard tokens is correct.
However, due to monotonically down-weighting all training examples, the improvement brought by \tfl{} is limited.

\begin{figure}
  \centering
  \includegraphics[width=.9\linewidth]{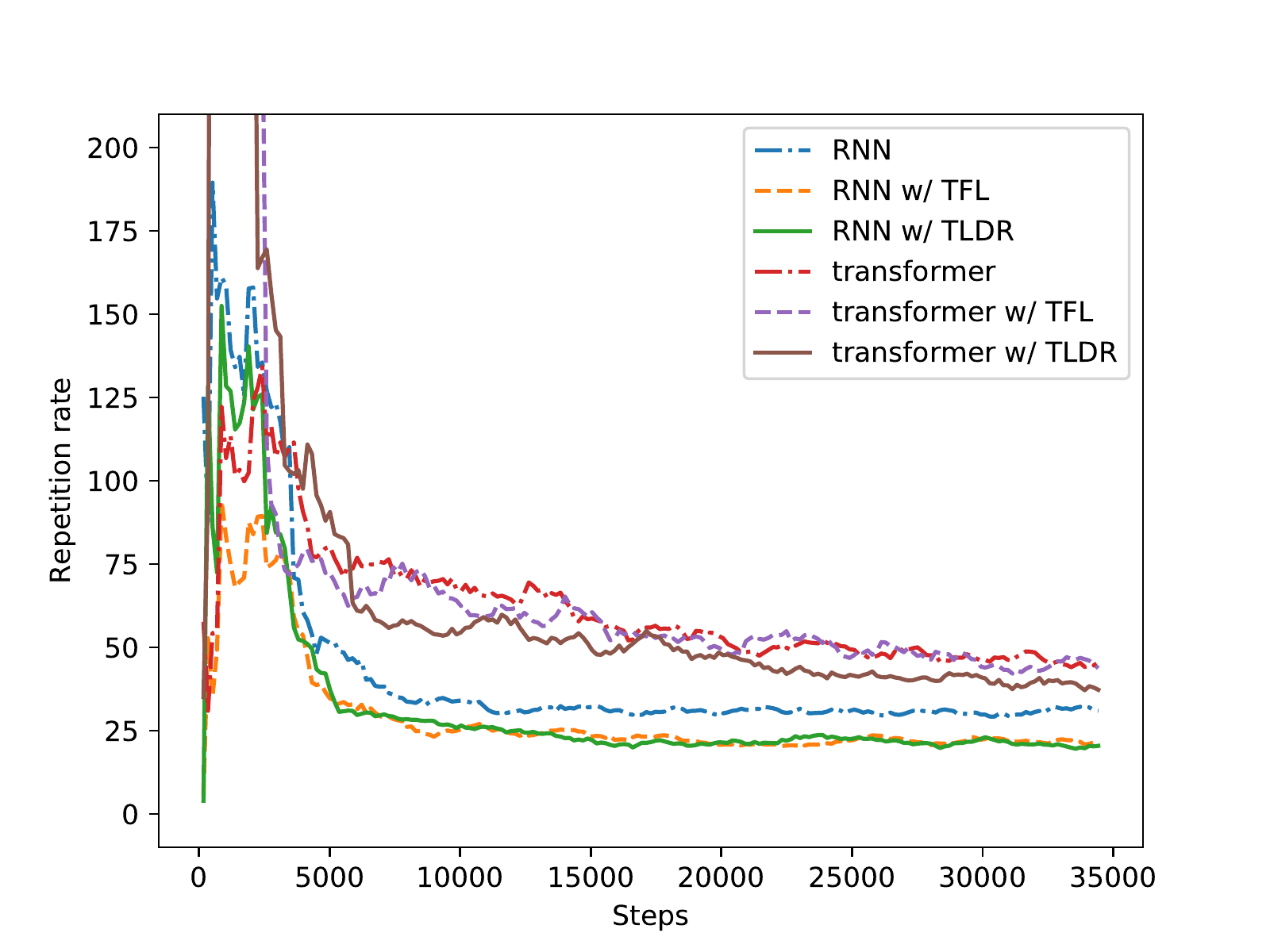}
  \caption{Repetition curves for Movie Dialogs validation set during training.
    TLDR can reduce the repetition rate earlier than baseline.
    Better viewed in color.}
  \label{fig:cornell_valid}
\end{figure}

\textbf{Answer to RQ3:}
Next, we proceed to our proposed weighting methods.
Giving that the LDR and TLDR methods are about model learning, thus should be architecture insensitive, we use the RNN variant (b) for the following experiments, as (c) needs more computation due to the utilization of a gating \eqref{eq:highway}.
From the second and third blocks of Table \ref{tab:results}, we can see that both RNN (row (g)) and transformer (row (l)) models trained using TLDR are consistently better than their corresponding baselines without weighting or with TFL weighting.
Though TLDR occasionally performs worse than TFL, e.g., row (g) vs row (e) on the Movie Dialogs dataset, we note that this is probably due to sample bias of the test sets.
We plot the result curves for the Movie Dialogs validation set at each validation point in Figure~\ref{fig:cornell_valid}, where we can see that TLDR is no worse, if not consistently better, than the baselines.
These comparisons indicate that dynamically weighting tokens according to their difficulty is an effective solution to reducing repetition.

The results in rows (d) and (i) show that, on the smaller datasets Dailydialog and Movie Dialogs, LDR has a limited effect on reducing repetition.
However, on the large Ubuntu Dialogue dataset, LDR hurts the repetition performance.
Besides, LDR consistently results in lower BLEU scores, which implies that more work needs to be done to estimate the training difficulty at the example level.

\textbf{Answer to RQ4:}
To understand whether the improvement of TLDR is brought by up-weighting, let us compare the performance of models trained using uniform weight $w = 2$ at rows (f) and (k) to rows (g) and (l), respectively.
We can see that w.r.t.\ repetition, TLDR wins the majority of the times, which supports our claim that TLDR has more effect than simply up-weighting, as can also be seen from the gradient visualization in Figure~\ref{fig:gradients}.
Multiplying a constant factor ($w = 2$ in our experiments) to the loss function also multiplies the gradients with the same factor, while TLDR changes the gradient function in a more complex way so that the gradients resulting from hard examples are amplified while those resulting from easy examples are suppressed, which is important for reducing repetition.
It is also worth noting that using a uniform weight of 2 is effectively doubling the base learning rate, which seems to be helpful with improving BLEU and \ldimen{}.




\section{Conclusion and Discussion}
\label{section:conclusion}

We have studied the repetition problem of encoder-decoder architectures, using both RNN and transformer models.
We have discovered that by using pre-attention instead of post-attention for \acp{RNN}, the repetition problem can be alleviated.
Together with highway connections, repetitive generations can be further reduced.

We have hypothesized that the repetition problem is caused by hard tokens and find empirical support for this claim using FL.
We then propose a more effective weighting function than FL, namely TLDR.
With a differentiable cosine weighting function, TLDR amplifies the gradients resulting from hard tokens while it suppresses those from easy tokens.
Through experiments we show that TLDR outperforms strong baselines like TFL and uniform weighting.

Our hard token hypothesis and the TLDR weighting function both operate mainly on target side tokens, while hard source side tokens can also have a detrimental effect on the decoder generations.
Future work can be done by applying TLDR to source side language representation learning, possibly by training the encoder independently on a language representation task.
It might also be worth exploring the effect on decoder-only models on tasks like \ac{LM}.

We also observe that in our experiments, before the repetition rate converges during training, the repetition rate on the Ubuntu Dialogues validation set fluctuates with a regular pattern, which suggests that there can be certain training examples that can harm the repetition rate.
If this is true, then the work by \citet{sharchilev2018finding} can be used to target such harmful examples and hence reduce repetition.
We leave this for future work.


\section*{Acknowledgments}

This research was supported by the China Scholarship Council, Ahold Delhaize, the Association of Universities in the Netherlands (VSNU), and the Innovation Center for Artificial Intelligence (ICAI).
All content represents the opinion of the authors, which is not necessarily shared or endorsed by their respective employers and/or sponsors.

\bibliographystyle{acl_natbib}
\bibliography{repetition-reduction}


\end{document}